\documentclass[alpha-refs, twocolumn]{wiley-article}

\usepackage[english]{babel}  
\usepackage{amssymb}  

\usepackage{url}  
\usepackage{hyperref}  
\hypersetup{colorlinks=true, allcolors=black, urlcolor=black}  

\usepackage{setspace}  

\usepackage{float}  

\usepackage{tikz}  
\newcommand{\blue}{\textcolor{black}}  

\papertype{Research Article}

\paperfield{Preprint Version}

\title{Exploring Wilderness Characteristics Using Explainable Machine Learning in Satellite Imagery}

\author[1]{Timo~T.~Stomberg}
\author[2]{Taylor~Stone}
\author[1,3]{Johannes~Leonhardt}
\author[4]{Immanuel~Weber}
\author[1,3]{Ribana~Roscher}

\affil[1]{Institute of Geodesy and Geoinformation, University of Bonn, Bonn, Germany}
\affil[2]{Institute for Science and Ethics, University of Bonn, Bonn, Germany}
\affil[3]{Data Science in Earth Observation, Technical University of Munich, Ottobrunn, Germany}
\affil[4]{AB|EX Community, PLEdoc GmbH, Essen, Germany} 

\corraddress{Timo T. Stomberg, Institute of Geodesy and Geoinformation, University of Bonn, Bonn, Germany}
\corremail{timo.stomberg@uni-bonn.de}

\runningauthor{\textbf{Preprint Version} of Exploring Wilderness Using Explainable Machine Learning in Satellite Imagery}  

\fundinginfo{German Federal Ministry for the Environment, Nature Conservation and Nuclear Safety (Grant Number: 67KI2043, KISTE); German Federal Ministry of Education and Research (Grant Number: 01DD20001, AI4EO); Alexander von Humboldt Foundation; Deutsche Forschungsgemeinschaft (Grant Numbers: EXC-2070 - 390732324, PhenoRob, and RO~4839/5-1 / SCHM~3322/4-1, MapInWild)}
    
\abbrevs{ASOS: Activation Space Occlusion Sensitivity, IIOS: Input Image Occlusion Sensitivity, ML: machine learning, WDPA: World Database on Protected Areas}

\begin{document}

\begin{frontmatter}
\maketitle

\begin{abstract}
Wilderness areas offer important ecological and social benefits and there are urgent reasons to discover where their positive characteristics and ecological functions are present and able to flourish.
We apply a novel explainable machine learning technique to satellite images which show wild and anthropogenic areas in Fennoscandia. Occluding certain activations in an interpretable artificial neural network we complete a comprehensive sensitivity analysis regarding wild and anthropogenic characteristics.
Our approach advances explainable machine learning for remote sensing, offers opportunities for comprehensive analyses of existing wilderness, and has practical relevance for conservation efforts.

\keywords{Activation Space Occlusion Sensitivity (ASOS),
AnthroProtect dataset,
concept discovery,
conservation,
deep learning,
explainable machine learning,
Sentinel-2,
wilderness
}

\end{abstract}
\end{frontmatter}

\begin{figure*}
    \centering
    \includegraphics[width=0.99\textwidth]{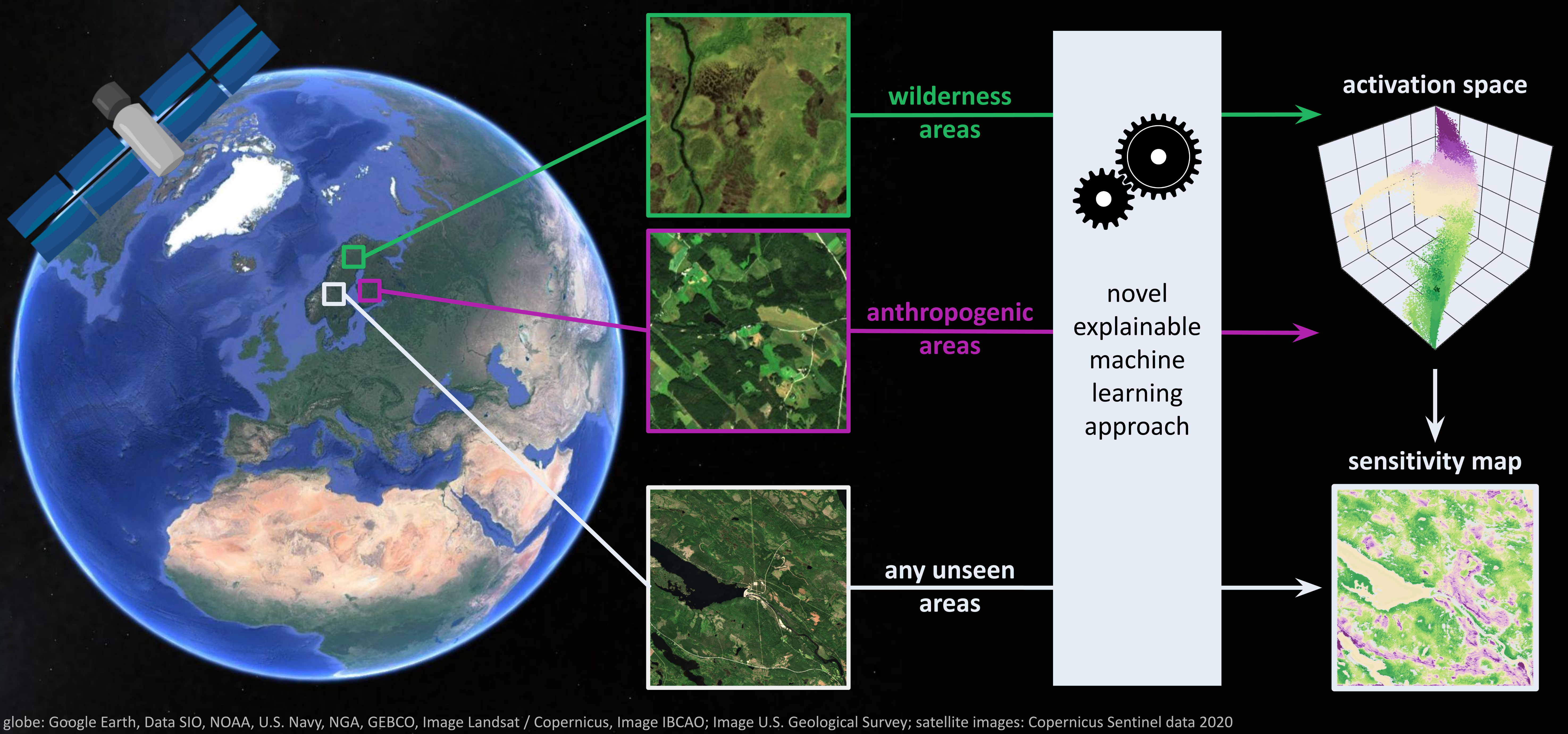}
\end{figure*}



\section{Introduction} \label{sec:introduction}

Within a very short period, from a geophysical point of view, humans have greatly expanded and they have strongly influenced Earth's environment \citep{steffen_anthropocene_2011}. Areas without human pressure on naturalness have been greatly reduced and hardly exist in most regions of the world \citep{allan_temporally_2017}. While urbanization and agriculture have brought many benefits, land use has had immense ecological impacts. However, countless species, including us, depend on natural ecosystem functions. Water cycles provide freshwater, forests regulate air quality, moors sequester carbon dioxide, pollinators are essential for the survival of flowering plants and successful harvests, etc. Disturbing natural ecosystems has an impact on biodiversity, pathogen spread, climate, and much more. The global consequences of human land use are well described by \cite{foley_global_2005}. Concerning this matter, wilderness areas can offer important ecological and social benefits; and there are urgent and pragmatic reasons (conservation, sustainable development, etc.) to identify where these positive characteristics and ecological functions associated with wilderness are present and able to flourish.

Using satellite imagery allows for continuous monitoring of broad, undisturbed regions without the need of visiting them. The Copernicus program and the USGS National Land Imaging Program provide free and open access to data of the satellites Sentinel-2 and Landsat~8, respectively, which are suitable for monitoring vegetation. Further, machine learning (ML) models are suitable for processing large amounts of data and can find patterns and relations that are potentially not recognizable by humans. \cite{ma_deep_2019} review on the variety of remote sensing applications in deep learning and \cite{roscher_explainable_2020} present the usefulness of such models being interpretable and explainable. Methods for explanations in deep learning are reviewed by \cite{samek_explaining_2021}.

A common explainable ML outcome are saliency maps which highlight regions that are important for the model's decision. \cite{kakogeorgiou_evaluating_2021} evaluated several methods for predicting saliency maps on models for land cover classification using Sentinel data and concluded that occlusion sensitivity maps by \cite{zeiler_visualizing_2014}, Grad-CAM by \cite{selvaraju_grad-cam_2017} and LIME by \cite{ribeiro_why_2016} lead to the most reliable results. \cite{zeiler_visualizing_2014} occlude patches in the input image and identify changes in the model's outcome. These changes are a measure of the occluded areas' sensitivities in regards to the observed class.

\paragraph{Our Approach and Objectives}

We build an artificial neural network that is interpretable by design allowing us to occlude certain activations in an interpretable layer and to measure the occurring deviations. In this way, we map the activations to sensitivities and predict detailed and high-resolution sensitivity maps. Analyses of the activation space increase confidence in the method, which builds upon the idea of \cite{stomberg_jungle-net_2021}.

The neural network is trained using Sentinel-2 images in protected and anthropogenic areas in Fennoscandia. Both categories are associated with the positive characteristics of wilderness and the unnatural influence of humans, respectively. Applying our sensitivity analysis, we inspect those areas and detect characteristics of wilderness in unseen regions in Fennoscandia.

Being able to analyze the extent and characteristics of wilderness is a valuable tool and can allow for the monitoring and tracking of conservation areas, reforestation efforts, etc. As such, it could be utilized as an indicator for policy-making and land-use planning, e.g. for the Sustainable Development Goal 15 of the \citeauthor{unitednations_sustainable_2021} (\citeyear{unitednations_sustainable_2021}, p.~56), which is focused on protecting, restoring, and promoting sustainable land use.

\section{A Philosophical Reflection on the Concept of Wilderness} \label{sec:wilderness}

It is generally agreed that wilderness areas offer important ecological and social benefits, and therefore warrant preservation and monitoring. Yet, the idea of wilderness is ethically and politically contested. It has been argued that wilderness is best understood as socially constructed: ``[W]hat constitutes wilderness is not the specific biophysical properties of an area but rather the specific meanings ascribed to it according to cultural patterns of interpretation'' (\citeauthor{kirchhoff_historical_2014}, \citeyear{kirchhoff_historical_2014}, p.~444). This has led to an ongoing debate regarding the role of wilderness in environmental decision-making and conservation efforts, captured in the edited volumes by \cite{callicott_great_1998} and \cite{nelson_wilderness_2008}. The primary critiques of wilderness are conceptual, targeting the meanings and associations it supports. These include the positioning of wilderness as places free of humans - and therefore sites of more ``authentic nature'' - thus reinforcing a dualism between humans and nature, and ignoring the historical presence of aboriginal peoples. For in-depth discussions of these critiques, see for example \cite{callicott_wilderness_1998}, \cite{cronon_trouble_1995} and \cite{vogel_thinking_2015}.

Given the global impact of human activities, the notion of sites with pristine nature completely unmodified by human actions, especially in Europe, is tenuous at best. We, therefore, do not posit that current wilderness areas are completely free of human presence or intervention. \citeauthor{alterra_guidelines_2013} (\citeyear{alterra_guidelines_2013}, p.~10) defines wilderness as ``an area governed by natural processes. It is composed of native habitats and species, and large enough for the effective ecological functioning of natural processes. It is unmodified or only slightly modified and without intrusive or extractive human activity, settlements, infrastructure or visual disturbance.'' Acknowledging this means we are not searching for some idealized or romanticized notion of ``authentic'' or ``true'' nature. Instead, we are searching for places in which the qualities of wilderness as defined above are best preserved and promoted. This may even include an active role for humans as managers to preserve natural conditions. Conversely, there are areas of pervasive and continuous human influence, which are actively and intentionally maintained for specific human purposes or functions (e.g., cities and communities, energy infrastructures, agriculture, etc.). This creates a categorically different type of human presence and intervention, which we here denote as ``anthropogenic''.

Another important clarification is the relation between ``wild'' and ``anthropogenic''. \cite{alterra_guidelines_2013} propose that wilderness is a spectrum according to the intensity of human interference. Seeing wild-to-anthropogenic spaces on a continuum can help to avoid binary thinking, instead appreciating there are degrees and types of anthropogenic influence. \cite{dill_defense_2021} points out, that certain extremes are in this sense anthropogenic or minimally wild (e.g., urban metropolitan cores). Conversely, other spaces (e.g., old-growth forests) exemplify the characteristics we would denote as wild or minimally anthropogenic.

While there is undoubtedly a need to be cautious with the label of ``wilderness'' due to historical contingencies and political concerns, there are undoubtedly places where, at the least, certain processes and biophysical conditions are better able to thrive. These positive characteristics are what we aim to discover.

\section{AnthroProtect - Satellite Imagery in Fennoscanida} \label{sec:dataset}

In searching for areas that align with our conceptualization of wilderness on the European continent, we identify certain protected areas within Fennoscandia, or more specifically the countries Norway, Sweden, and Finland, as suitable. We are aware that over the last 300 years the landscapes of Fennoscandia, namely forests, have seen anthropogenically-driven changes before regulations were introduced to protect them. This affects the southern regions more than the northern ones. \cite{kouki_forest_2001} and \cite{ostlund_history_1997} give detailed overviews of forest fragmentation in Fennoscandia and the transformation of the boreal forest landscape in Scandinavia, respectively. However, as discussed in Section~\ref{sec:wilderness}, we are not searching for idealized nature and appreciate the longstanding, strict conservation efforts in certain protected regions. All three countries have high environmental standards according to the Environmental Performance Index by \cite{wendling_2020_2020}. Furthermore, both the human influence index by \cite{sanderson_human_2002} and the wilderness quality index by \cite{fisher_review_2010} map wide areas within Fennoscandia as areas with relatively minimal (disruptive) anthropogenic influence.

On this basis, we build the AnthroProtect dataset consisting of Sentinel-2 images in Fennoscandia of 1)~protected areas to preserve and retain the natural character and 2)~anthropogenic areas consisting of artificial and agricultural landscapes. For this dataset, we do not intend to make assumptions about wilderness continuity as described in Section~\ref{sec:wilderness} but leave this task to our ML model. Instead, we look for areas that are at the extreme ends of the continuity scale, which we then categorize into two groups: ``wild'' and ``anthropogenic''.

The workflow for data extraction is mainly based on Google Earth Engine by \cite{gorelick_google_2017}. The full dataset and the code for the data export are available at \url{http://rs.ipb.uni-bonn.de/data/anthroprotect} and \url{https://gitlab.jsc.fz-juelich.de/kiste/asos}, respectively.

\paragraph{Finding Regions of Interest}
In order to find areas that are associated with wilderness, we use the World Database on Protected Areas (WDPA) by \cite{unep-wcmc_protected_2021} which contains polygons of protected areas according to the categories of \cite{dudley_guidelines_2008}. We consider terrestrial areas of categories Ia (strict nature reserve), Ib (wilderness area), and II (national park) with a minimum area of 50~km\textsuperscript{2}. The corresponding class is called ``wild'' in the following.

To find areas that are not associated with wilderness, we use the Copernicus CORINE Land Cover dataset by the \cite{europeanenvironmentagency_corine_2018}. First, we locate areas with land cover classes 1 (artificial surfaces) and 2 (agricultural areas). Then, three morphological functions are applied to these areas in the following order: 1)~closing to remove holes and gaps, 2)~opening to increase compactness and filter small structures, and 3)~dilation to create a buffer. For closing and opening, we use a circle with a radius of 2 km, respectively, and for dilation a circle with a radius of 1 km. Finally, all areas are filtered for a minimum area of 50~km\textsuperscript{2}. We call the corresponding class ``anthropogenic'' in the following.

\paragraph{Multispectral Sentinel-2 Data}
Within Fennoscandia, we assume vegetation to provide significant indicators for the appearance of wilderness. Therefore, we use multispectral imagery of the Sentinel-2 satellites whose instruments are specialized in vegetation. We decide to use Sentinel-2 over Landsat 8, due to the better spatial resolution and the additional red-edge bands. Further, \cite{astola_comparison_2019} concluded in a study that Sentinel-2 outperforms Landsat~8 in predicting forest parameters in Finland.

\begin{figure}
    \centering
    \includegraphics[width=0.9\columnwidth]{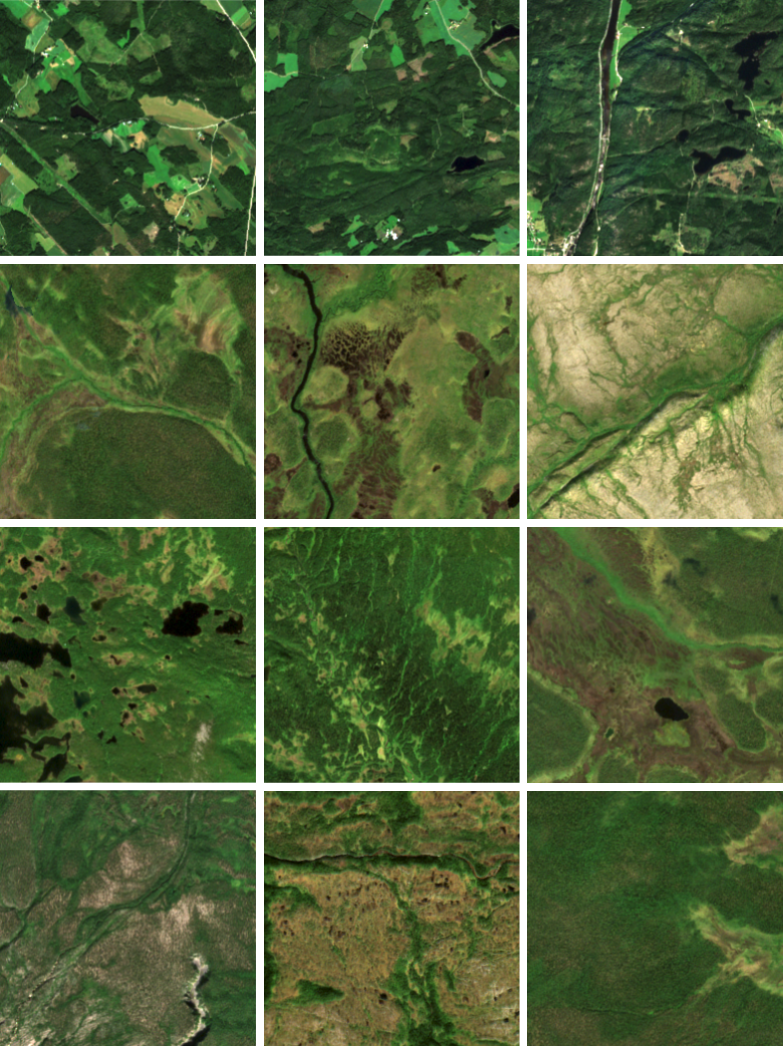}
    \caption{Samples of Sentinel-2 images. Shown are three samples for each category. From top to bottom: 1)~anthropogenic, 2)~WDPA Ia, 3)~WDPA Ib, 4)~WDPA II. \tiny{Copernicus Sentinel data 2020.} \normalsize{}}
    \label{fig:dataset_samples}
\end{figure}

For each wild and anthropogenic area, we export one cloud-free, multi-temporal Sentinel-2 image composite. Each image composite is calculated from a multi-temporal image collection. For each region, the workflow is as follows:
1)~The atmospheric corrected Sentinel-2 products (Level-2A) are used with a resolution of 10~meters.
2)~Images are filtered for the time period of summer 2020 (July 1\textsuperscript{st} to August 30\textsuperscript{th}).
3)~A mask for clouds, cirrus, and cloud shadows is created for each image using the Quality-60~m band (QA60) and scene classification map (SCL) provided by Sentinel-2. Only images with a mask fraction of less than 5~\% within the region of interest are taken.
4)~The masked areas are dilated with a radius of 100~meters to prevent artifacts at the transitions.
5)~For each pixel and band, the 25\textsuperscript{th} percentile is calculated across all remaining images in that region. Masked areas are not taken into account. This way, we receive a single image composite for that region.
6)~The following ten bands are exported: B2, B3, B4, B5, B6, B7, B8, B8A, B11, B12.
7)~We look at the red-green-blue channels (B4, B3, B2) in person and remove the image composite if it has strong artifacts. This concerns a total of two regions in the whole dataset.
8)~Each region of interest is tiled into images of size 256~$\times$~256~pixels, which corresponds to 2560~$\times$~2560~meters.
Samples for each category are shown in Figure~\ref{fig:dataset_samples}.

\paragraph{Land Cover Data}
Besides Sentinel-2 images, the following land cover data is exported for each region, that we use for evaluation purposes only:
1)~a composite (most common value) of the Sentinel-2 scene classification map (SCL),
2)~the Copernicus CORINE Land Cover dataset by the \cite{europeanenvironmentagency_corine_2018},
3)~the MODIS Land Cover Type 1 by \cite{friedl_mcd12q1_2019},
4)~the Copernicus Global Land Service by \cite{buchhorn_copernicus_2020} and
5)~the ESA GlobCover by \cite{arino_global_2012}.
The distribution of the CORINE Land Cover classes over the AnthroProtect dataset is shown in Table~\ref{tab:lc_classes}

\begin{table*}
    \centering
    \begin{tabular}{lrr}
        CORINE Land Cover class                            & wild / \% & anthropogenic / \% \\ \hline
        11 urban fabric                                    &       0.0 &                2.4 \\
        21 arable land                                     &       0.0 &               26.4 \\
        23 pastures                                        &       0.0 &                1.0 \\
        24 heterogeneous agricultural areas                &       0.0 &                8.2 \\
        31 forest                                          &      34.6 &               50.8 \\
        32 shrub and/or herbaceous vegetation associations &      27.1 &                3.7 \\
        33 open spaces with little or no vegetation        &      16.9 &                0.3 \\
        41 inland wetlands                                 &      14.6 &                1.1 \\
        51 inland waters                                   &       3.8 &                5.0 \\
        other                                              &       3.0 &                1.1 \\
    \end{tabular}
    \caption{\label{tab:lc_classes}Distribution of the CORINE Land Cover classes in the AnthroProtect dataset separated by classes. \normalsize{}}
\end{table*}

\paragraph{Data Split}
The data is divided into three subsets for training, validation, and testing. The three subsets are intended to be independent of each other, spatially consistent, and categorically consistent. To ensure categorical consistency, the data split is performed separately for each category (WDPA categories Ia, Ib and II, and anthropogenic). To ensure independence and spatial consistency, spatial clusters are built as follows: In the first step, data samples are separated if their distance is larger than 10 km using the clustering algorithm DBSCAN developed by \cite{ester_density-based_1996}. This way, some large clusters occur, so that in the second step, these large clusters are spatially clustered using the k-means algorithm by \cite{lloyd_least_1982}. We use scikit-learn by \cite{pedregosa_scikit-learn_2011} to perform both clustering algorithms.

Subsequently, all samples within one cluster are assigned to the same dataset. We choose the split fractions to be \blue{80~\%~/~10~\%~/~10~\%}. Hereby, samples within very small clusters are assigned to the training dataset.

\begin{figure}
    \centering
    \includegraphics[width=0.9\columnwidth]{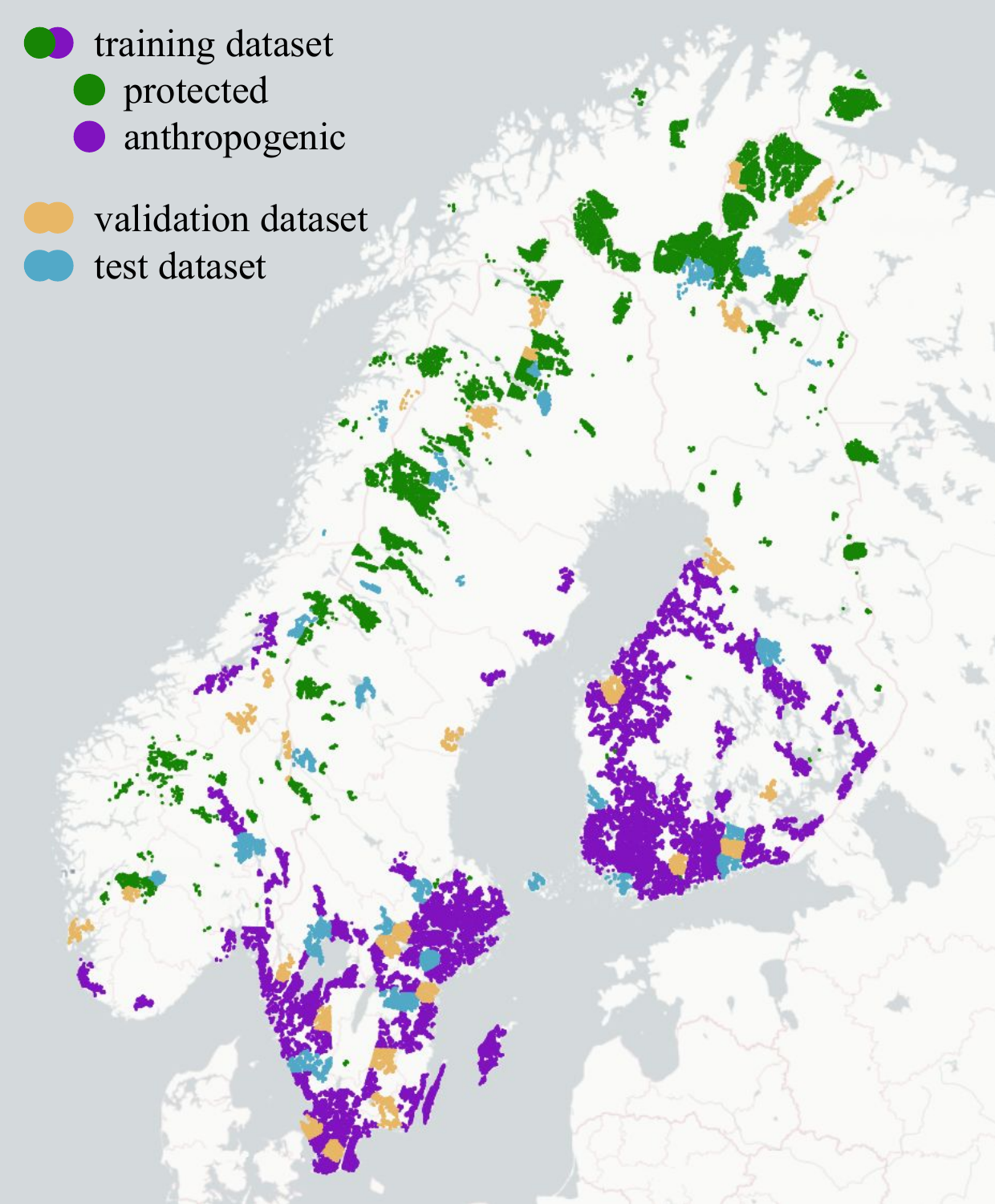}
    \caption{Locations of the AnthroProtect data samples. Shown are the \blue{7,003} wild and \blue{16,916} anthropogenic samples which are split into three independent subsets for training (\blue{80~\%}), validation (\blue{10~\%}) and testing (\blue{10~\%}). For better clarity, the coloring of both classes differs only for the training set. \tiny{The plot is created with \cite{plotlytechnologiesinc_collaborative_2015} and copyright holders of the map are Carto and OpenStreetMap contributors.} \normalsize{}}
    \label{fig:dataset_locations}
\end{figure}

Our procedure of data splitting, compared with a random data split, reduces the incidence that nearby samples appear in different datasets. This prevents validation and test results from being glossed over. The resulting data split is visualized in Figure~\ref{fig:dataset_locations} and the sizes of each subset and category are listed in Table~\ref{tab:dataset}.

\begin{table*}
    \centering
    \begin{tabular}{llrrr|r}
        class         & WDPA category & \# train & \# val & \# test & \# total \\ \hline
        wild          & Ia            &      295 &     37 &      37 &      369 \\
                      & Ib            &    3,601 &    465 &     446 &    4,512 \\
                      & II            &    1,693 &    220 &     209 &    2,122 \\
        anthropogenic & -             &   13,534 &  1,670 &   1,712 &   16,916 \\ \hline
                      &               &   19,123 &  2,392 &   2,404 &   \blue{23,919}
    \end{tabular}
    \caption{\label{tab:dataset}Number of samples in the AnthroProtect dataset separated by categories and subsets. \normalsize{}}
\end{table*}

\paragraph{Investigative Areas}
Besides the mentioned wild and anthropogenic regions, some further Sentinel-2 images are exported for regions that are of interest for investigation. This includes several villages, forests, power plants, wind parks, airports, and more. For many of these regions, time series of the years 2017 to 2021 are included in this dataset. \blue{It is ensured that all investigative regions do not overlap with samples of the training, validation, or test set.}

\section{Activation Space Occlusion Sensitivity} \label{sec:methodology}

Our presented methodology, Activation Space Occlusion Sensitivity (ASOS), is a two-step procedure: First, an artificial neural network - consisting of an encoder-decoder network and a classifier - is trained to classify images. After training the whole model, the activation maps at the interface of the two neural networks are analyzed: Certain activations are occluded causing deviations in the classification score. In doing so, we map these activations to sensitivity values in terms of the classification decision. Having a trained neural network and a functional relationship between activations and sensitivities, we can predict sensitivity maps of any input images. Our method builds upon the idea of analyzing activation maps within an activation space as suggested by \cite{stomberg_jungle-net_2021}. The code for the here presented methodology and experiments is available at \url{https://gitlab.jsc.fz-juelich.de/kiste/asos}.

\begin{figure*}
    \centering
    \includegraphics[width=0.99\textwidth]{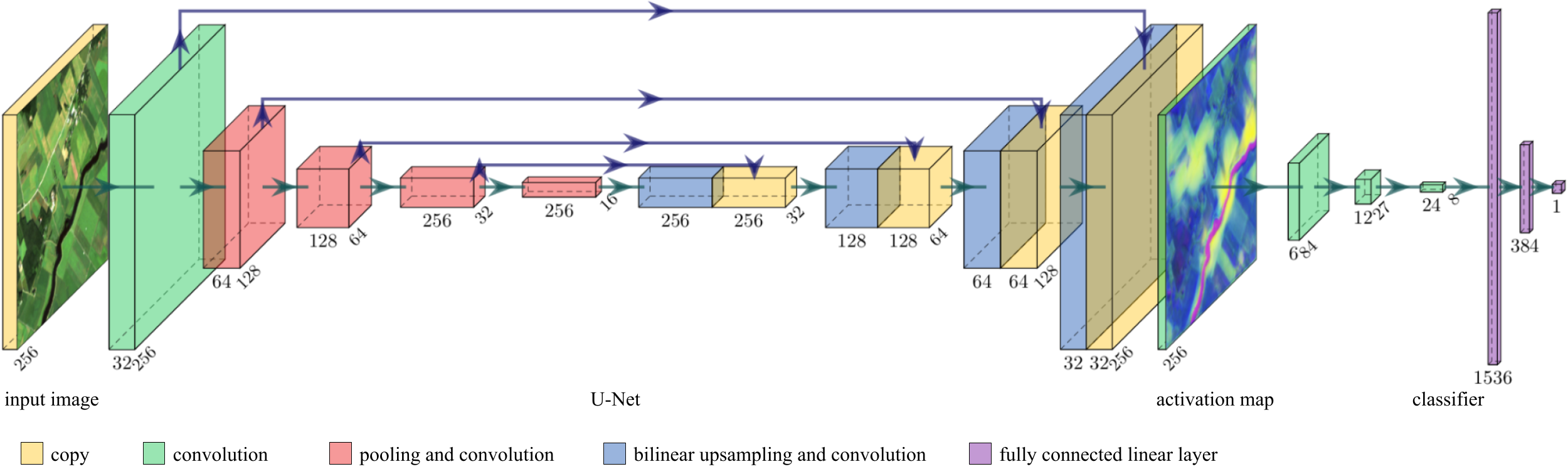}
    \caption{Architecture of the artificial neural network. It consists of a modified U-Net, followed by an image classifier. It takes multispectral images and predicts a score $\hat{y} \in [0, 1]$. The activation map at the interface between both networks has the same height $h$ and width $w$ as the input image. \tiny{The figure has been created with PlotNeuralNet by \cite{iqbal_harisiqbal88plotneuralnet_2018}. On the left-hand side it contains a Sentinel-2 image of the Copernicus Sentinel data 2020.} \normalsize{}}
    \label{fig:nn_architecture}
\end{figure*}

\paragraph{Neural Network Architecture}
For the experiments shown in this article we use 1)~a modified form of the U-Net by \cite{navab_u-net_2015} and 2)~a classifier with significantly fewer parameters consisting of convolutional and linear layers. Our architecture is visualized in Figure~\ref{fig:nn_architecture}. Both U-Net and classifier are treated as a single neural network during training and are therefore trained end-to-end.

Our modified U-Net has the following characteristics:
1)~It consists of four encoding and four decoding steps. Instead of two convolutional layers per encoding or decoding step, our U-Net has only one such layer. Furthermore, we reduce the number of output channels for each convolution as shown in Figure~\ref{fig:nn_architecture}.
2)~We add batch normalization after each convolutional layer.
3)~Including padding to each convolution, we preserve the image size at each skip connection.
4)~We replace the deconvolutional upsampling with bilinear upsampling as proposed in \cite{odena_deconvolution_2016} to prevent checkerboard artifacts.
5)~Instead of a single-channel input image, our U-Net takes $n_\textrm{in}$-channel input images. Furthermore, we consider the number of activation map channels as a hyperparameter~$n_\textrm{m}$. The activation map is the output of the U-Net.
6)~The activation map is not batch-normalized and activated with the hyperbolic tangent function (tanh) so that the activation map has values in the range of~-1~and~1.
With this architecture, the predicted activation map of an image with shape ($h$,~$w$,~$n_\textrm{in}$) has shape ($h$,~$w$,~$n_\textrm{m}$)~-~so height~$h$ and width~$w$ remain unchanged.

The classifier network consists of three convolutional layers followed by two fully connected linear layers. Each convolutional layer doubles the number of channels, has a kernel size of 5, a stride of 3, and is ReLU activated. The output of the last convolutional layer is flattened. Two fully connected linear layers follow with 128 and 1 neuron(s), respectively. The first linear layers are ReLU activated and the last one is activated with Sigmoid, yielding predictions $\hat{y} \in [0, 1]$ in the range of 0 (anthropogenic) to 1 (wild).

\begin{figure*}
    \centering
    \includegraphics[width=0.99\textwidth]{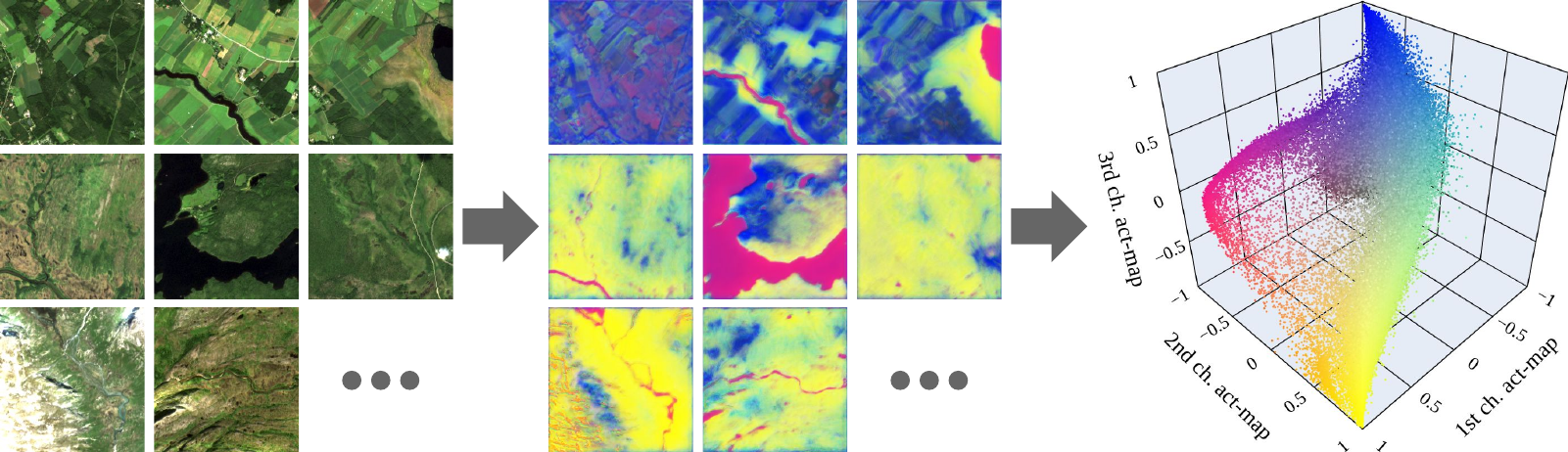}
    \caption{Activation maps and their representation within the activation space. The left-hand side shows samples of the \blue{19,074} correctly classified Sentinel-2 images of the training dataset. For each of them, one activation map is predicted with the encoder-decoder network. Here, each activation map has $n_\textrm{m} = 3$ channels so that they can be displayed as red-green-blue images in the middle of this figure. Each activation (pixel) in all activation maps is represented within the three-dimensional activation space (right-hand side), in which each axis represents the values of one of the three channels. \tiny{The images on the left-hand side are Sentinel-2 images of the Copernicus Sentinel data 2020. The plot on the right-hand side is created with \cite{plotlytechnologiesinc_collaborative_2015}.} \normalsize{}}
    \label{fig:activation_space_rgb}
\end{figure*}

\paragraph{Activation Space}
After the neural network has been trained, we analyze the activation space, which we define as follows: Having $N$ correctly classified training samples, we obtain $N$ activation maps at the interface between both networks. We treat each activation (pixel) in each activation map as a vector representing the $n_\textrm{m}$ channels. These activations build an $n_\textrm{m}$-dimensional activation space, in which each axis represents the values of one of the $n_\textrm{m}$ channels. The described steps are visualized in Figure~\ref{fig:activation_space_rgb} for the specific case $n_\textrm{m} = 3$.

\begin{figure}
    \centering
    \includegraphics[width=0.8\columnwidth]{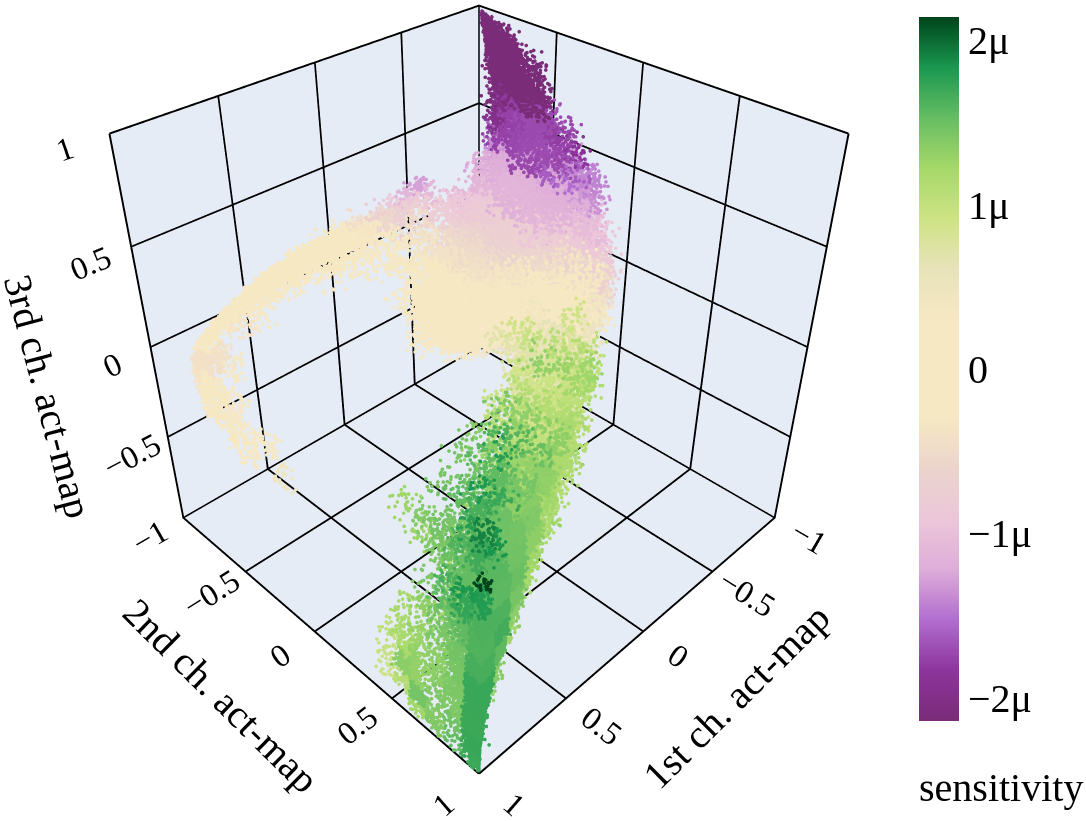}
    \caption{Sensitivities in the activation space. The greener the activations are represented, the more sensitive they are to wild characteristics; the more purple, the more sensitive they are to anthropogenic characteristics; the more beige, the less sensitive they are. For low-density areas, the sensitivity is not determined which is why the corresponding activations are not displayed here. The minimum and maximum values of the color scale are chosen according to the \blue{2~\%}-percentile of the absolutes of all existing values. In the colorbar, $\mu$ stands for 1e-6. \tiny{The plot is created with \cite{plotlytechnologiesinc_collaborative_2015}.} \normalsize{}}
    \label{fig:activation_space_sensitivities}
\end{figure}

\paragraph{Occlusions and Sensitivities}
We map areas in the activation space to sensitivities as illustrated in Figure~\ref{fig:activation_space_sensitivities}. To determine these sensitivities, we occlude activations in the activation maps that are close within the activation space. These occlusions lead to deviations in the classification scores and these are a measure of the sensitivity of the occluded regions. In detail, this works as follows:

We slide a $n_\textrm{m}$-dimensional hypercube with edge length $l_\textrm{cube}$ and stride $l_\textrm{cube}$ through the activation space. If the density of activations within the hypercube is higher than a given threshold, we perform the following steps:
1)~We occlude all activations within the hypercube in the activation maps by setting the actual values to zero. This is only done for all correctly classified training samples.
2)~We pass the partially occluded activation maps to the classifier and receive a prediction $\hat{y}_{\textrm{occ}}$ for each activation map.
3)~Comparing each prediction with the corresponding non-occlusion prediction $\hat{y}$, we obtain a deviation per activation of $\delta = (\hat{y}_{\textrm{occ}} - \hat{y}) / n_{\textrm{occ}}$, where $n_{\textrm{occ}}$ is the number of occluded activations in the corresponding activation map. In doing so, we obtain one deviation for each of the $N$ correctly classified training samples.
4)~We define a threshold and consider only those deviations where $n_\textrm{occ}$ is larger. In this way, we skip very small deviations due to very few occluded activations.
5)~We define the negative of the mean value over all remaining deviations to be a measure of the sensitivity within the hypercube: $\eta = - \langle \{ \delta \} \rangle$ where $\{ \delta \}$ is the set of all deviations for the activations within the hypercube.

\paragraph{Sensitivity Maps}
Having a trained encoder-decoder network and determined activation space sensitivities, we can predict sensitivity maps for any images. First, the activation map is determined using the trained encoder-decoder network. Second, this activation map is evaluated using the functional relationship between activation values and sensitivity, as derived from the activation space occlusions. Since our encoder-decoder network is a purely convolutional neural network, images can have any height~$h$ and width~$w$. Samples are shown in \blue{Figures~\ref{fig:inv_letsi}, \ref{fig:inv_alvdal} and \ref{fig:inv_forestry}}.

Sensitivities were not determined for low-density regions within the activation space of the training samples. Therefore, activation maps might have activations that cannot be attributed to a sensitivity value. This characteristic prevents the sensitivity map to be filled when the sensitivity is unlikely to be correct. We mask these areas using the color grey. 

\paragraph{CutMix and Loss Function}
We determine sensitivities by occluding certain areas in the activation maps after the model has been trained. For this purpose, the neural network must be sensitive to small changes. However, when training the model with hard labels only (0 and 1), we observe that the sensitivity scores have distinct peaks at values towards 0 and 1. Implementing CutMix similar to \cite{yun_cutmix_2019} during training leads to a more balanced allocation of sensitivity scores.

\begin{figure}
    \centering
    \includegraphics[width=0.99\columnwidth]{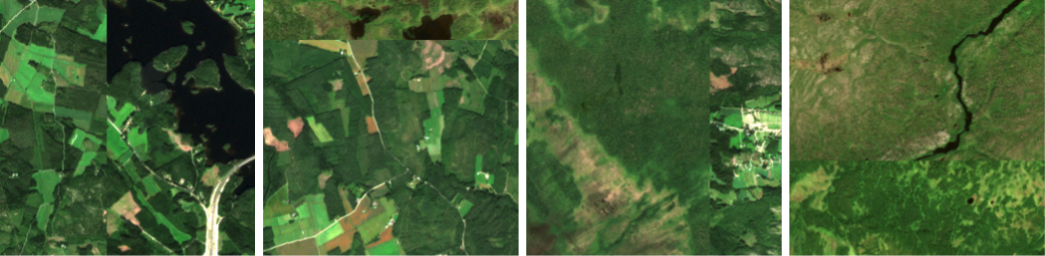}
    \caption{Cut-Mix samples. From left to right: 1)~two anthropogenic areas are Cut-Mixed resulting in label~0; 2)~an anthropogenic and a wild area are Cut-Mixed resulting in label~0.16; 3)~a wild and an anthropogenic area are Cut-Mixed resulting in label~0.72; 4)~two wild areas are Cut-Mixed resulting in label~1. \tiny{Copernicus Sentinel data 2020.} \normalsize{}}
    \label{fig:cutmix}
\end{figure}

We apply CutMix on the input images with a chance of 80~\% while training the model. If CutMix is applied, a stripe is cut from another random training sample and pasted at a random edge with a random amount of size between 0 and 50~\%. The label is adjusted proportionally which may result in a continuous value. Samples of resulting images and their labels are shown in Figure~\ref{fig:cutmix}.

\cite{yun_cutmix_2019} use CutMix as a data augmentation technique with a cross-entropy loss function. We, on the other hand, use CutMix to increase the model's ability to predict continuous scores between 0 and 1 and therefore use mean square error loss. Having trained the model with this technique, it can predict even low deviations, e.g. if only small areas are occluded.

\paragraph{Neutral Occlusion Value}
\cite{sturmfels_visualizing_2020} point out the challenge of making features missing in neural networks. In terms of our method, the related question is which value to use when replacing the values of activations that are intended to be occluded. We deliberately use tanh to activate the activation maps, because it ranges from -1 to 1 which makes value~0 a good choice to deactivate certain features. In addition, we randomly occlude activations in the activation maps by setting them to this value with a chance of 20~\% to 50~\% while training the model. This way, we simulate occlusions during training so that the classifier learns to treat activations with the value zero as neutral.

\paragraph{Comparison Method}
Determining sensitivities by occlusions is based on the idea by \cite{zeiler_visualizing_2014}. Therefore, we choose their approach as a comparison method and produce Input Image Occlusion Sensitivity (IIOS) maps as follows:

We use the same, trained model as for ASOS to ensure comparability. Instead of the activation maps, we occlude the input images. Occlusions are not defined by the activation space, but by a square-shaped patch with edge length $l_\textrm{patch}$. We slide this patch through the input image with a stride of $l_\textrm{stride}$. For each position, we perform the following steps:
1)~We replace the values of each pixel covered by the patch with zero.
2)~We pass the occluded input image to the model and receive a prediction of $\hat{y}_\textrm{occ}$.
3)~Comparing each prediction with the corresponding non-occlusion prediction $\hat{y}$, we obtain a deviation per pixel of $\delta = (\hat{y}_\textrm{occ} - \hat{y}) / (l_\textrm{patch})^2$.
This way, we receive one deviation for each occlusion. There exist overlapping occlusions, if the stride $l_\textrm{stride}$ is lower than the edge length $l_\textrm{patch}$ of the patch. In this case, the mean value of the deviations is calculated and we define the negative of that value to be a measure of the sensitivity.

\section{Experiments} \label{sec:experiments}

\paragraph{Model and Training Setup}
We build our model using PyTorch by \cite{paszke_pytorch_2019}. According to the multi-spectral Sentinel-2 data, the number of input channels is set to $n_\textrm{in} = 10$. Further, we choose the number of activation map channels to be $n_\textrm{m} = 3$. Other choices are possible as discussed in Section~\ref{sec:discussion}. Overall, our modified U-Net has about 1.8 million parameters. The classifier is significantly smaller with about 200,000 parameters.

The Sentinel-2 Level-2A products have values from 0 to 10,000. We scale all values to be in a range from 0 to 1. Additionally to CutMix, we perform random image rotations of 90°, 180°, or 270° during training to increase variability. Our model is trained with a batch size of~32. We optimize the model's parameters with stochastic gradient descent and perform the one cycle learning rate policy by \cite{smith_super-convergence_2019} with a maximum learning rate of \blue{1e-2}. Further, we add a weight decay of \blue{1e-4} to the loss function. In total, the model is trained for \blue{5} epochs on a \blue{NVIDIA Quadro RTX 4000 (8 GB GDDR6)} for about \blue{15 minutes}.

\paragraph{Accuracy Assessment}
The achieved overall accuracies of the training, validation and test data are \blue{99.7~\%}, \blue{99.96~\%} and \blue{99.7~\%}, respectively. Table~\ref{tab:confusion_matrix} shows the confusion matrix of the test dataset.

\begin{table}
    \centering
    \begin{tabular}{l|rr}
                      & prediction:   &           \\
                      & anthropogenic & wild      \\ \hline
        label:        &               &           \\
        anthropogenic &         1,710 &         2 \\
        wild          &             4 &       688         
    \end{tabular}
    \caption{\label{tab:confusion_matrix}Confusion matrix of the test dataset which has an accuracy of \blue{99.7~\%}. \normalsize{}}
\end{table}

\paragraph{Activation Space}
With our trained model, we predict $N = \blue{19,074}$ activation maps for all correctly classified training samples and analyze them within the activation space as described in Section~\ref{sec:methodology}. Because of border effects in the activation maps due to padding layers in the U-Net, we define a small frame size of $\blue{10}$ pixels and do not consider activations at the margins within this frame. Furthermore, we only use a randomly chosen fraction of \blue{1e-3} of all activations. The activation space, as well as some activation maps, are visualized in Figure~\ref{fig:activation_space_rgb}.

\begin{figure}
    \centering
    \includegraphics[width=0.99\columnwidth]{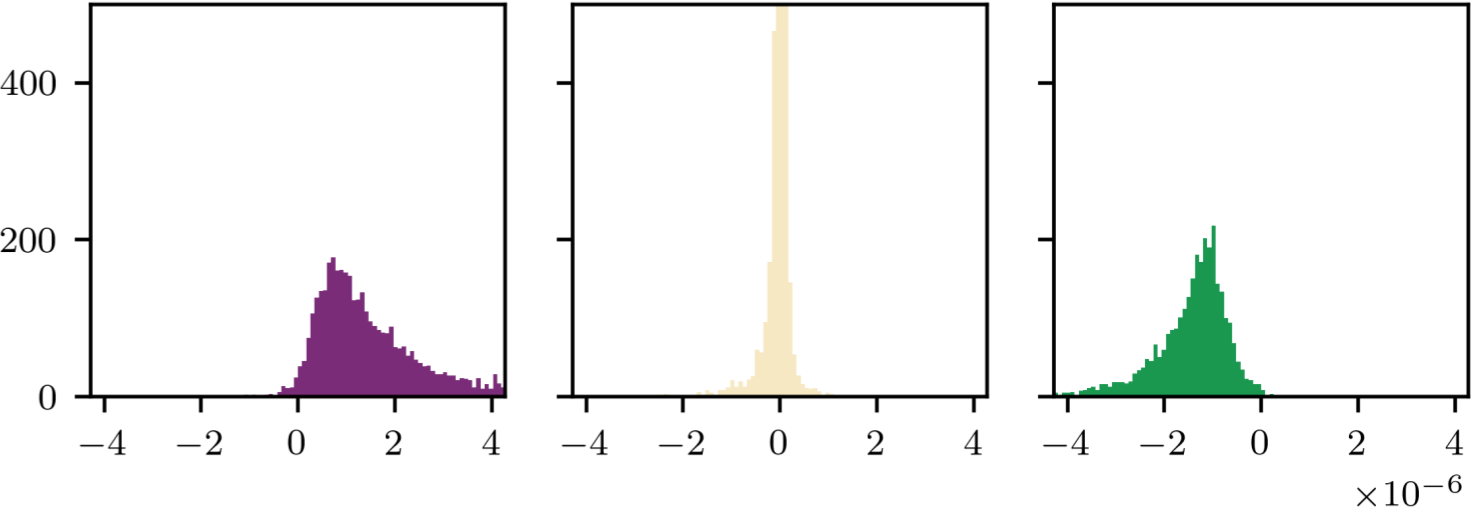}
    \caption{Deviations due to occlusions of activations for three different positions of the hypercube. From left to right:
    1)~Occluding activations in the activation maps that are within the hypercube mainly causes positive deviations. The sensitivity is $\eta = - \langle \{ \delta \} \rangle = - 2.0 \times 10^{-6}$ which means that the area in the activation space is sensitive to anthropogenic characteristics. 2)~The deviations are distributed around zero and so is the sensitivity. The corresponding area in the activation space is neither sensitive to anthropogenic nor wild characteristics. 3)~Most deviations are negative and the sensitivity is $1.8 \times 10^{-6}$ which means sensitive to wild characteristics. \tiny{The plot is created with Matplotlib by \cite{hunter_matplotlib_2007}.} \normalsize{}}
    \label{fig:deviation_histograms}
\end{figure}

\paragraph{Occlusions and Sensitivities}
Determining the sensitivities, we decide the edge length of the hypercube to be $l_\textrm{cube} = \blue{0.1}$ and choose a minimum density threshold of \blue{two} times the average density. Again, we ignore the margins of the activation maps by not occluding activations within the defined frame size of $\blue{10}$. We define $\blue{10}$ to be the threshold for the minimum number of occluded activations. Figure~\ref{fig:deviation_histograms} shows deviations due to occluding activations within the same hypercube. We take the negative of the mean value of the deviations as a measure of sensitivity. The resulting activation space sensitivities are visualized in Figure~\ref{fig:activation_space_sensitivities}.

\begin{figure}
    \centering
    \includegraphics[width=0.99\columnwidth]{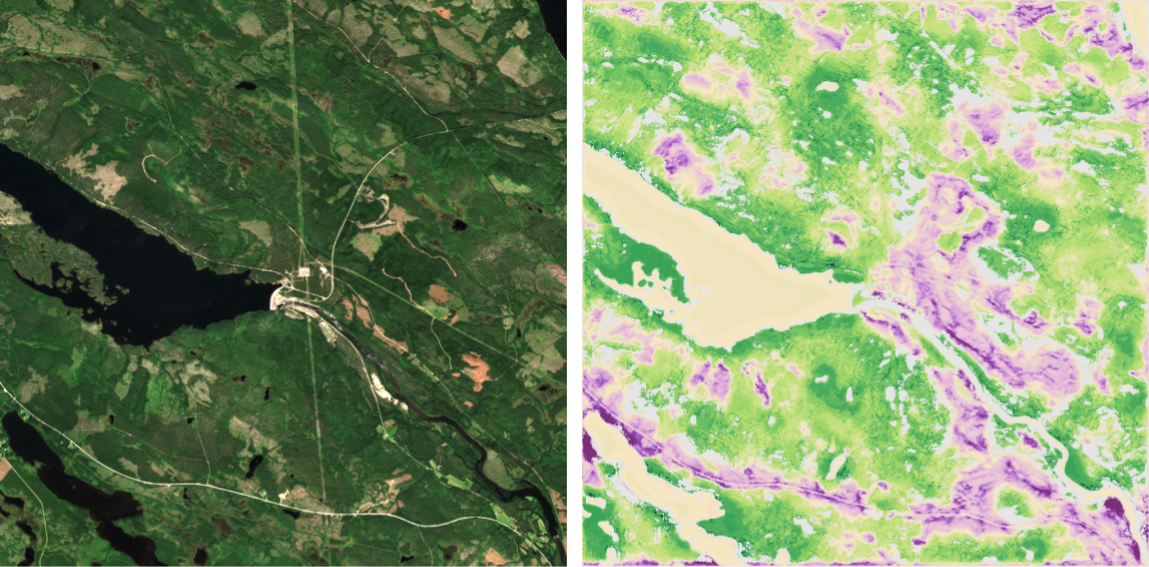}
    \caption{Hydroelectric power plant Letsi in Sweden, located in an area barely inhabited by humans. Shown are a Sentinel-2 image and its sensitivity map covering an area of about 100 km\textsuperscript{2}. The color scale is the same as in Figure~\ref{fig:activation_space_sensitivities}. Sensitivities not predicted due to a low density in the activation space are colored in grey. \tiny{Copernicus Sentinel data 2020.} \normalsize{}}
    \label{fig:inv_letsi}
\end{figure}

\begin{figure}
    \centering
    \includegraphics[width=0.99\columnwidth]{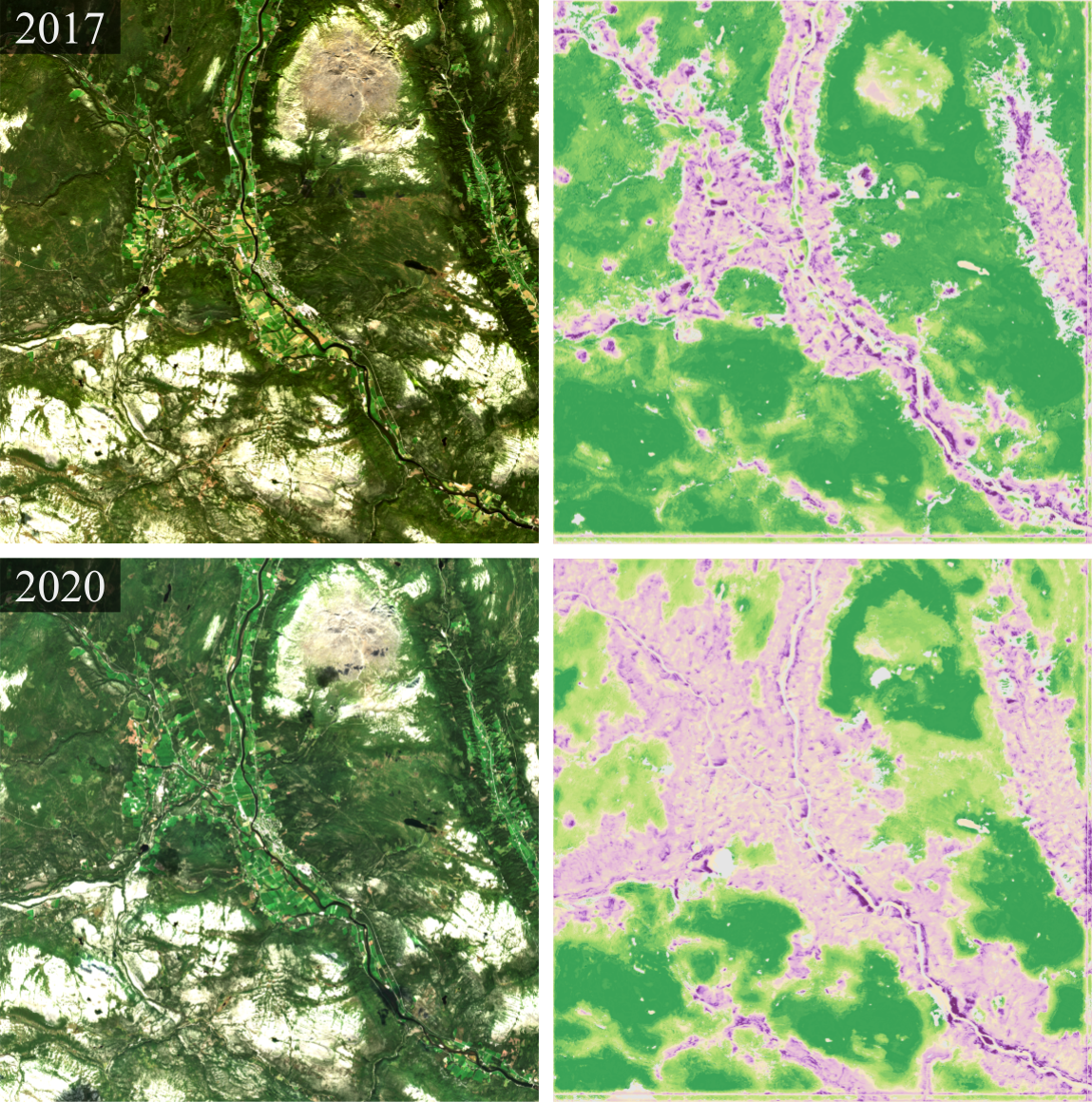}
    \caption{Municipality Alvdal in Norway, whose villages are located within the Østerdalen valley. Shown are Sentinel-2 images and their sensitivity maps covering an area of about 420 km\textsuperscript{2} each. Deforestation within the years from 2017 to 2020 causes the model to predict a larger anthropogenic area in 2020. The color scale is the same as in Figure~\ref{fig:activation_space_sensitivities}. Sensitivities not predicted due to a low density in the activation space are colored in grey. \tiny{Copernicus Sentinel data 2017 and 2020.} \normalsize{}}
    \label{fig:inv_alvdal}
\end{figure}

\begin{figure}
    \centering
    \includegraphics[width=0.99\columnwidth]{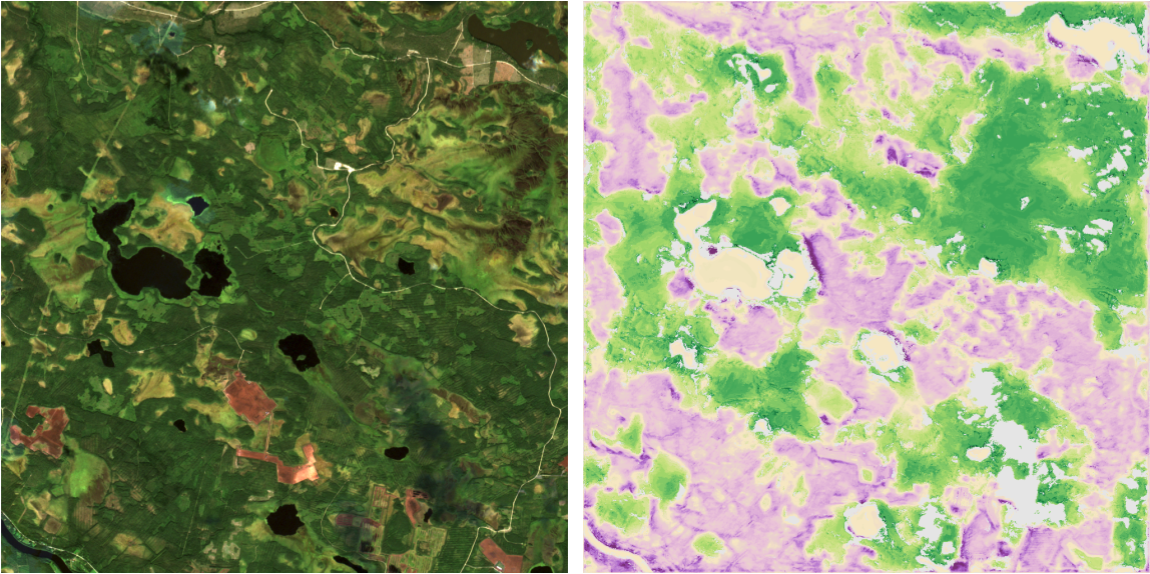}
    \caption{Forestry and unused wetlands encounter in North Ostrobothnia, a region of Finland. Shown are a Sentinel-2 image and its sensitivity map covering an area of about 100 km\textsuperscript{2}. The color scale is the same as in Figure~\ref{fig:activation_space_sensitivities}. Sensitivities not predicted due to a low density in the activation space are colored in grey. \tiny{Copernicus Sentinel data 2020.} \normalsize{}}
    \label{fig:inv_forestry}
\end{figure}

\paragraph{Sensitivity Maps}
With our trained model and the determined activation space sensitivities, we predict sensitivity maps for some of the investigative areas included in the AnthroProtect dataset shown in \blue{Figure~\ref{fig:inv_letsi}, \ref{fig:inv_alvdal} and \ref{fig:inv_forestry}.}

\begin{figure}
    \centering
    \includegraphics[width=0.9\columnwidth]{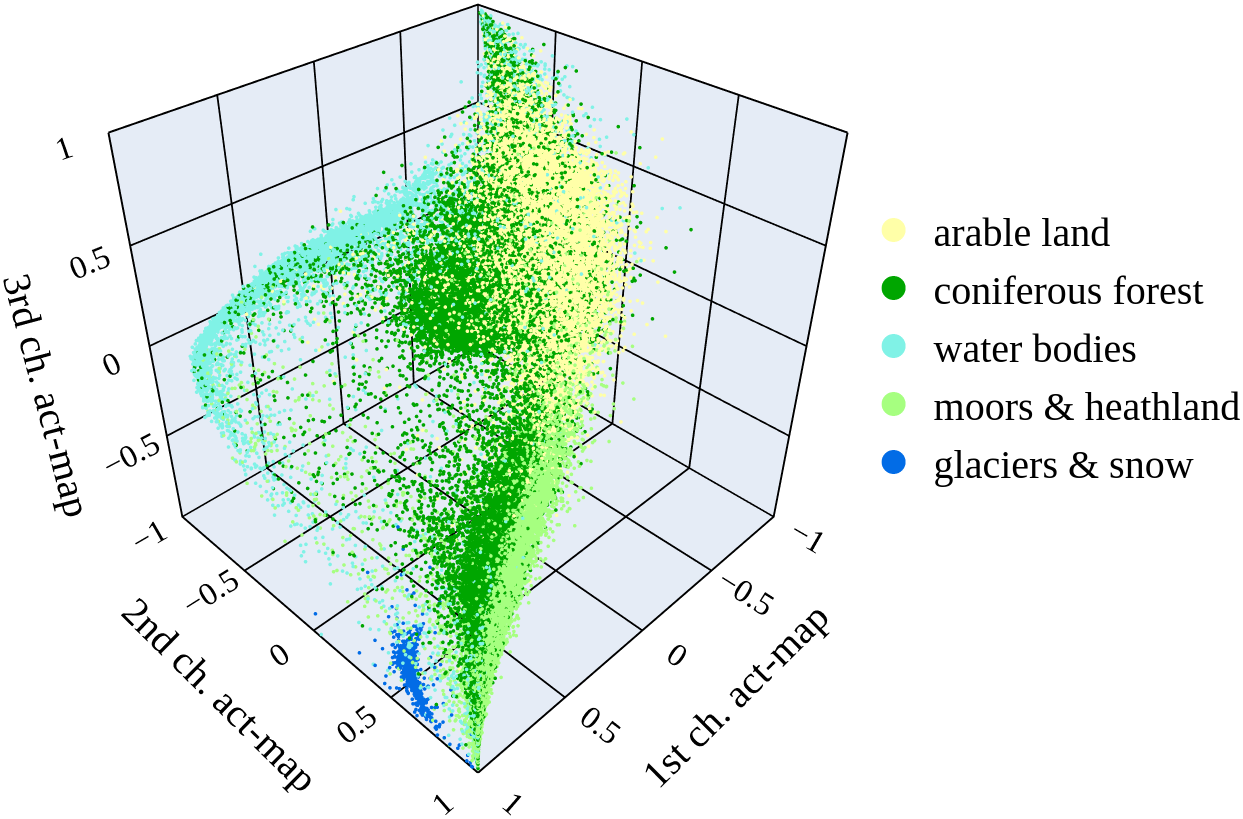}
    \caption{Land cover classes within the activation space. Shown is a subset of classes of the Copernicus CORINE Land Cover dataset by the \cite{europeanenvironmentagency_corine_2018}. Here, the land cover names are shortened and the color of the class glaciers \& snow has been changed for a clearer visualization. From top to bottom, the official classes are 211, 312, 512, 322 and 335. \tiny{The plots are created with \cite{plotlytechnologiesinc_collaborative_2015}.} \normalsize{}}
    \label{fig:activation_space_lc}
\end{figure}

\paragraph{Land Cover Classes within the Activation Space}
The AnthroProtect dataset provides land cover data so that we can relate each activation to a land cover class. Figure~\ref{fig:activation_space_lc} shows a subset of the CORINE land cover classes within the activation space. A division of these classes can be well seen. We obtain similar results with the other land cover data available in the AnthroProtect dataset.

\begin{figure}
    \centering
    \includegraphics[width=0.99\columnwidth]{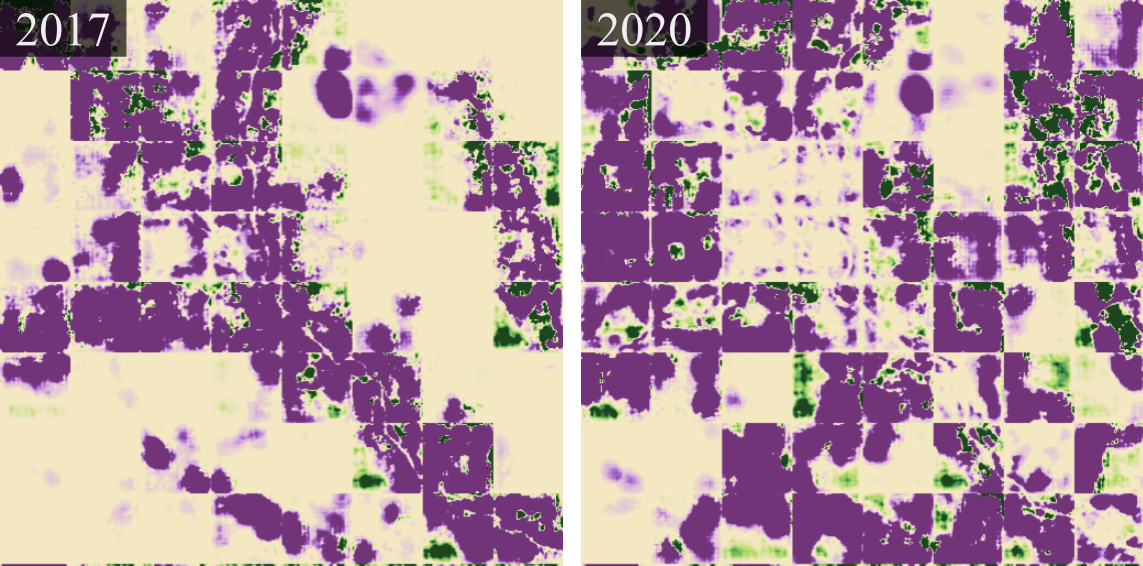}
    \caption{IIOS maps of Alvdal. Sentinel-2 images and ASOS maps are shown in Figure~\ref{fig:inv_alvdal}. The color scale is multiplied by \blue{10} compared to the ASOS maps. To predict the IIOS maps, the image is subdivided into tiles of size $256 \times 256$. \normalsize{}}
    \label{fig:iios}
\end{figure}

\paragraph{Comparison Method}
We predict IIOS maps using a patch's edge length of $l_\textrm{patch}$~=~\blue{8} and a stride of $l_\textrm{stride}$~=~\blue{4}. Other than in ASOS, we need the whole model including the classifier to predict IIOS for investigative samples. Since the classifier has linear layers, IIOS can be predicted only for images with a height and width of $h = w$~=~256 pixels. We, therefore, split the investigative images into tiles of this size and merge them after the prediction of the IIOS maps. Figure~\ref{fig:iios} shows the IIOS results for \blue{the Alvdal valley in Norway}. The corresponding ASOS results are shown in Figure~\ref{fig:inv_alvdal}.

\section{Discussion} \label{sec:discussion}

\subsection{Technical Discussion}

\paragraph{Accuracy Assessment}
Building the AnthroProtect dataset, we localized areas that are at the ends of the continuity scale between wild and anthropogenic (see Section~\ref{sec:dataset}). The high classification accuracies of the model ($>$99.7~\% for all subsets) shows that there exist clear decision characteristics between the two chosen classes. One advantage of having an interpretable model is, that we can rule out undesirable effects such as presented in \cite{lapuschkin_unmasking_2019}, which falsely drive the accuracy upwards.

\paragraph{Semantic Arranging within the Activation Space}
We make several observations supporting the hypothesis that regions are semantically arranged within the activation space.
1)~Negative and positive sensitivities are separated within the activation space and there is an approximately even gradient in between (see Figure~\ref{fig:activation_space_sensitivities}). This indicates that the model arranges regions based on their influence on the classification score.
2)~We observe that within the same hypercube, nearly all deviations are either positive or negative. Representative histograms are shown on the left- and right-hand sides of Figure~\ref{fig:deviation_histograms}. If the mean value is close to zero, the distribution is usually very narrow. It hardly happens that opposite deviations occur in the same hypercube.
3)~We find that regions are arranged according to specific land cover classes, as shown in Figure~\ref{fig:activation_space_lc}. Some of the classes tend to merge into each other, which is coherent considering that mixed landscapes are not uncommon. Also, faulty assignments may occur due to the resolution of the CORINE dataset of 100~m, which is significantly less than the resolution of the Sentinel-2 images of 10~m; and due to the accuracy of the CORINE dataset of about 92~\%.

\paragraph{Sensitivity Maps}
We observe sensitivity maps predicted with ASOS to be detailed and in high resolution. Structures as the lake in Figure~\ref{fig:inv_letsi} are mapped nearly pixel-wise and so are many fields in Figure~\ref{fig:inv_alvdal} and \ref{fig:inv_forestry}. In contrast, the resolution of IIOS maps is much more coarse (see Figure~\ref{fig:iios}), because of the given patch shape. This issue is not present with ASOS maps, because occluded activations can cover unshaped and disjoint areas. We further observe that ASOS maps are robust regarding the size of the hypercube. A larger hypercube does not directly influence the spatial resolution but only causes more step-like transitions between sensitivity values. Other than in IIOS, this does not change the essential appearance of the sensitivity maps, because of the semantical closeness of occluded activations.

Another advantage of ASOS maps is the comparability of sensitivity values. With ASOS, we do not determine sensitivities for a specific image but rather for groups of activations within the whole training dataset. This makes different sensitivity maps comparable to each other. This is not the case for IIOS maps, where sensitivities are independently determined for each image. The consequence is visible in Figure~\ref{fig:iios}, year 2020. Tiles amid the valley are less sensitive than tiles at the sides of the valley.

Once all parameters are set, the calculations of ASOS maps are computationally inexpensive. The input image passes the U-Net once, then the activation maps are transferred to sensitivity maps. The prediction of IIOS maps is much more computationally expensive since all occluded input images have to pass the whole neural network. On the upside, IIOS can be applied to any model whereas ASOS depends on a specific model architecture.

\paragraph{Hyperparameters}
We observe our method to be robust in regards to the choice of hyperparameters and model architecture. We obtain reasonable results with learning rates in the range of \blue{1e-3} to \blue{1e-1} and weight decays in the range of \blue{0} to \blue{1e-1}. The higher the weight decay, the denser the activations in the activation space. Also, batch normalization is important for a wide-spread activation space.

\paragraph{Number of Activation Map Channels}
We present results obtained with $n_\textrm{m} = 3$ activation map channels which results in a 3-dimensional activation space. However, one can decide on a different number of activation map channels. If $n_\textrm{m} = 2$, the activation space can be visualized on a plane and if $n_\textrm{m} = 1$, it can be visualized as a histogram. High dimensional activation spaces could be visualized using dimensionality reduction techniques. However, the number of possible hypercube positions exponentially grows with the number of dimensions, which makes the sensitivity analysis more computationally intensive.

Training a model with $n_\textrm{m} = 2$, we observe that the 2-dimensional activation space looks similar to the projection of the 3-dimensional activation space onto a plane. Also, the sensitivity maps are of equal quality. However, a consequence is that the land cover classes are no longer as well-separated in the 2-dimensional activation space.

\paragraph{Multi-Class Classification}
In our research, we have two classes and our classifier predicts a single confidence score between 0 and 1. However, our methodology could be generally applied to classifiers predicting a $n_\textrm{c}$-dimensional vector where $n_\textrm{c}$ is the number of classes. In that case, a sensitivity value would be also a vector with $n_\textrm{c}$ dimensions. Visualizing the activation space sensitivities, one would receive $n_\textrm{c}$ different visualizations - one for each class. Likewise, one would receive $n_\textrm{c}$ sensitivity maps for each image.

\paragraph{Applicability on Other Data}
ML models generally show a low generalization ability for data that has a significantly different distribution than the training data. Although the AnthroProtect dataset covers a variety of land cover classes (see Table~\ref{tab:lc_classes}), outside of Fennoscandia the presented model is likely not applicable - particularly if the region has distinctly different landscapes or ecosystems such as savannas or tropical forests. However, training ASOS on different regions of interest and/or different classes will also produce valuable results. To demonstrate the versatile applicability, we present experiments with photography images in Appendix~\ref{sec:places365}. In addition, we refer to the MapInWild dataset by \cite{ekim_mapinwild_2022}, which contains globally distributed samples of protected areas.

\subsection{Analyzing Investigative Samples}

The analysis of the following investigative samples is representative of many others. We observe similar results in other areas, especially in regards to water, deforestation, villages, agricultural areas, roads, power lines, edges, structures, and wetland areas.

\paragraph{Hydroelectric Power Plant Letsi in Sweden}
The hydroelectric power plant Letsi in Sweden (Figure~\ref{fig:inv_letsi}) is located in an area barely inhabited by humans.
1)~Areas close to the power plant (center) are mainly highlighted as anthropogenic in the sensitivity map.
2)~The water reservoir (left of the power plant) and the other lake (bottom left) do not seem to contribute to the classification decision. Matching the separation of the land cover class water in the activation space (Figure~\ref{fig:activation_space_lc}) with the corresponding sensitivities (Figure~\ref{fig:activation_space_sensitivities}), we see that water bodies are commonly neutral. The model does not distinguish between the unnatural water reservoir and the natural lake.
3)~No sensitivities could be attributed to the river (starting right of the power plant) due to a low density of activations within the activation space of the training data.
4)~Right above the river, we see deforestation areas. In the sensitivity map, they are much more extended. One deforestation area, however, is only highlighted in its surroundings. Here, not the area itself but the edges seem to trigger the model.
5)~The power line (going centered from top to bottom) is not detected as anthropogenic. In contrast, the road along the bottom is well highlighted in its surroundings; other roads are not or only barely.
6)~In general, dark violet highlighted areas are usually narrow and close to edges, whereas dark green highlighted areas tend to be larger and often occur centrally within areas sensitive to wilderness.

\paragraph{Municipality Alvdal in Norway}
The villages of the municipality Alvda in Norway are located within the Østerdalen valley shown in Figure~\ref{fig:inv_alvdal}.
1)~In the year 2017, we see that the model detects villages and agricultural areas as anthropogenic and all forests and rocky landscapes as wild.
2)~According to the global forest loss mapping by \citeauthor{hansen_high-resolution_2013} (\citeyear{hansen_high-resolution_2013}, \url{https://gfw.global/3qJLGPX}) there has been significant tree cover loss within the years 2017 to 2020. Deforestation areas can be also seen in the Sentinel-2 image of 2020. They cause the model to predict a much larger anthropogenic area in that year.
3)~In both images, we see that dark violet highlights are often textured whereas dark green highlights usually cover large areas.

\paragraph{North Ostrobothnia in Finland}
In North Ostrobothnia, a region in Finland, forestry and unused wetlands encounter (according to LUCAS data by \citeauthor{eurostat_lucas_2018}, \citeyear{eurostat_lucas_2018}) and build an interesting landscape shown in Figure~\ref{fig:inv_forestry}.
1)~Roads and a power line are not highlighted here.
2)~Remaining cloud shadows (bottom right) are grayed out.
3)~The huge wetland area (top right), as well as many of the smaller wetland areas, are sensitive to wilderness.
4)~On the other hand, most but not all forests are detected as anthropogenic.
5)~The brownish fields in the lower part of the image are peat production areas. Interestingly, these areas are wild for the model despite human peat mining. We believe that this misconception occurs because there are too few or no peat production areas within our training data.

\subsection{Conclusions in Regards to Wilderness}

\paragraph{Deforestation and Disruption Areas}
We observe that our model detects certain types of unnatural disruption, such as deforestation, even within small regions which in turn leads to fairly dramatic shifts in sensitivities. In doing so, the model can distinguish between disrupted and natural bare soils. It seems that not just the size but also the number of deforested areas has an impact on the sensitivity map, as in the municipality Alvdal (Figure~\ref{fig:inv_alvdal}). Here, many small areas become disproportionately sensitive. This behavior may also allow conclusions about a minimum contiguous wild-like area needed to be graded as actually wild.

\paragraph{Edges and Uniformity}
We observe that in most samples dark violet highlighted areas are narrow and close to edges whereas dark green highlighted areas tend to be larger and occur centrally within areas sensitive to wilderness. This suggests that specific edges seem to be important anthropogenic characteristics whereas scale and uniformity are important factors of wilderness.

\paragraph{Roads and Adjacent Vegetation}
We have reason to believe that our model is not sensitive to roads but to the surrounding vegetation. This assumption is based on the fact that both small and large roads, and both asphalted or gravel roads are sometimes recognized as anthropogenic and sometimes not. Furthermore, we observe other linear structures like power lines not to be highlighted by the model. Both roads and power lines are very narrow compared with the image resolution of 10~$\times$~10 meters per pixel. Vegetation next to roads could be affected by direct land use or indirect influences such as pollutants due to construction, traffic, or de-icing salt carried into surrounding soils. The ecological effects of roads and traffic are well reviewed by \cite{spellerberg_ecological_1998}. If we do not misinterpret the sensitivity maps here, they show that some roads can strongly influence the adjacent vegetation.

\paragraph{Type of Human Influence}
Forestry is conducted in North Ostrobothnia (Figure~\ref{fig:inv_forestry}) as well as near the power plant Letsi (Figure~\ref{fig:inv_letsi}) according to the LUCAS data by \cite{eurostat_lucas_2018}. However, we observe that trees in North Ostrobothnia are planted in distinct rows, different from trees near Letsi. Human influence seems to be of a different type in regards to the preservation of naturalness and the model recognizes these differences in landscape management.

\paragraph{Limitations}
One limitation of our method is that we cannot detect anthropogenic influence on wildlife unless the affected wildlife influences the vegetation in a measurable amount. For example, habitat fragmentation due to roads is a particular problem for many species of wildlife. Literature examples are listed by \citeauthor{spellerberg_ecological_1998} (\citeyear{spellerberg_ecological_1998}, Table~2). Furthermore, plants or soil obscured by larger plants cannot be detected using satellite images. This especially appears in forests where tree canopies hide mosses and grasses etc. Human influence must affect the appearance of tree canopies or we cannot detect such influences.

\section{Conclusion} \label{sec:conclusion}

In this article, we explore the characteristics of wilderness using satellite images and explainable ML. Hereby, we refer less to the absence of humans but more to the aims and goals with which an area is managed and to what extent natural ecological functions are present and able to flourish. Based on our AnthroProtect dataset, we predict sensitivity maps highlighting wild and anthropogenic characteristics in Fennoscandia applying the explainable ML technique ASOS. Our model distinguishes between different types of human influence and detects altered vegetation adjacent to roads. It further differentiates between disrupted and natural bare soils and is therefore capable of detecting deforestation areas. As such, our approach has the potential to be used for monitoring and evaluating conservation efforts using present satellite data and offers opportunities for comprehensive analyses of existing wilderness.


\section*{Data and Code Availability}
The AnthroProtect dataset is available at \url{http://rs.ipb.uni-bonn.de/data/anthroprotect}. The code for AnthroProtect's data export, the presented methodology ASOS and the presented experiments is available at \url{https://gitlab.jsc.fz-juelich.de/kiste/asos}.

\section*{Acknowledgements}
We thank Michael Schmitt and Burak Ekim from the University of the Bundeswehr Munich for the idea of investigating wilderness, the valuable discussions, and their support with the creation of the AnthroProtect dataset. We acknowledge funding from the German Federal Ministry for the Environment, Nature Conservation and Nuclear Safety under grant no 67KI2043 (KISTE), the German Federal Ministry of Education and Research (BMBF) in the framework of the international future AI lab ``AI4EO -- Artificial Intelligence for Earth Observation: Reasoning, Uncertainties, Ethics and Beyond'' (Grant number: 01DD20001) and the Alexander von Humboldt Foundation in the framework of the Alexander von Humboldt Professorship endowed by the Federal Ministry of Education and Research. This work has partially been funded by the Deutsche Forschungsgemeinschaft (DFG, German Research Foundation) under Germany's Excellence Strategy, EXC-2070 - 390732324 - PhenoRob. In addition, we acknowledge funding from DFG as part of the project RO~4839/5-1 / SCHM~3322/4-1 - MapInWild.

\section*{Declaration of Interests}
The authors declare that they have no known competing financial interests or personal relationships that could have appeared to influence the work reported in this article.

\section*{Author Contributions}
All authors conceived the ideas. Timo T. Stomberg led the writing of the manuscript. Taylor Stone focused on the conceptual questions of wilderness. Timo T. Stomberg and Johannes Leonhardt collected the data. Timo T. Stomberg, Ribana Roscher, Immanuel Weber and Johannes Leonhardt designed the methodology.  Timo T. Stomberg ran the experiments. All authors analyzed the results. All authors contributed critically to the drafts and gave final approval for publication.

\appendix

\section{Places365 Dataset} \label{sec:places365}

To show that ASOS is not limited to satellite imagery, we apply it to chosen classes of the Places365-Standard dataset by \cite{zhou_places_2018}. They provide a small image version with 256~$\times$~256 pixels in size, which is equal to the Sentinel-2 images' size of the AnthroProtect dataset. We use the same model architecture and only change the number of input channels $n_\textrm{in}$ to 3 instead of 10 since the Places365 images have red, green, and blue channels only. Again, we scale the images from 0~to~1, perform CutMix, rotation augmentation, and random occlusions during training and use the same hyperparameters. We increase the number of epochs to \blue{20} though since we have fewer data samples than in AnthroProtect.

\begin{figure}
    \centering
    \includegraphics[width=0.99\columnwidth]{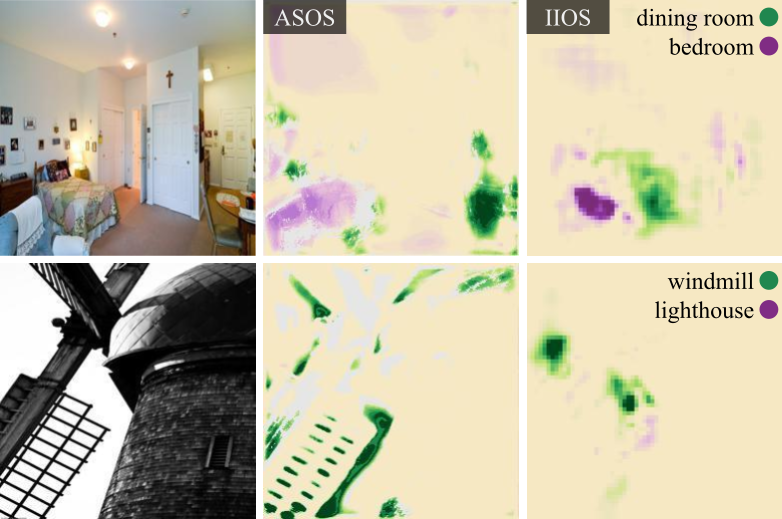}
    \caption{Experiments with the Places365 dataset by \cite{zhou_places_2018}. Upper row: Sensitivity maps resulting from training with the classes dining room and bedroom. Lower row: Sensitivity maps resulting from training with the classes windmill and lighthouse. Sensitivities not predicted due to a low density in the activation space are colored in grey in the ASOS maps. \tiny{The plot is created with Matplotlib by \cite{hunter_matplotlib_2007}.} \normalsize{}}
    \label{fig:places365}
\end{figure}

For our experiments we use the classes 1)~dining room vs. bedroom and 2)~windmill vs. lighthouse. For each of the classes, the Places365-Standard dataset provides \blue{5,000} training images and \blue{100} validation images. We randomly split the training set into two subsets so that, for each class, we get \blue{4,900} training samples and \blue{100} test samples. For each of the two experiments, we predict the activation space sensitivities using all correctly classified training samples and apply them to our test data. Furthermore, we predict IIOS maps for the test data. Samples are shown in Figure~\ref{fig:places365}.

ASOS works meaningfully on the Places365 dataset. Compared with IIOS, the same advantages stand out as described in Section~\ref{sec:discussion}. Comparing the ASOS maps of both datasets, we observe two main differences. 1)~ASOS maps often highlight edges in the Places365 images as with the windmill blade in Figure~\ref{fig:places365}, whereas in the AnthroProtect images areas are highlighted more often. 2)~In Places365 images, the sensitivity maps are mainly neutral, whereas in the AnthroProtect dataset neutral areas occur more rarely. It seems that in photographic images edges are more characteristic for classification, whereas in remote sensing images areal structures are of more importance.

\bibliography{references}

%
%


\end{document}